%% file: iwslt2019.tex
\documentclass[a4paper]{article}
\usepackage{iwslt18,amssymb,amsmath,epsfig}
\def\reg{{\rm\ooalign{\hfil
     \raise.07ex\hbox{\scriptsize R}\hfil\crcr\mathhexbox20D}}}

\usepackage{url}

\usepackage{adjustbox}
\usepackage{amsmath}
\usepackage{bm}
\usepackage{booktabs}
\usepackage{url}
\usepackage{siunitx}
\usepackage{graphicx}
\usepackage[utf8x]{inputenc}
\usepackage{multirow}
\usepackage{enumitem}
\usepackage{longtable}
\usepackage{hyperref}

\usepackage{xcolor}

\providecommand{\citet}[1]{\cite{#1}}

\newcommand*{\metric}[1]{\num[round-mode=places,round-precision=1]{#1}}

\def\Snospace~{\S{}}

\usepackage[textwidth=0.7in]{todonotes}

\newcommand{\up}[1]{\textsuperscript{\textup{#1}}}

\title{Harnessing Indirect Training Data for End-to-End Automatic Speech Translation: \\ Tricks of the Trade}

\makeatletter
\def\name#1{\gdef\@name{#1\\}}
\makeatother
\name{{\em Juan Pino\up{1}, Liezl Puzon\up{1}, Jiatao Gu\up{1}, Xutai Ma\up{1,2}, Arya D. McCarthy\up{1,2}, Deepak Gopinath\up{1}}}

\address{\up{1}Facebook \\ \up{2}Johns Hopkins University \\
\texttt{\{juancarabina,lie,jgu,dgopinath\}@fb.com} \texttt{\{xutai\_\thinspace ma,arya\}@jhu.edu} 
}

\begin{document}
\maketitle

\begin{abstract}
    For automatic speech translation (AST), end-to-end approaches are outperformed by cascaded models that transcribe with automatic speech recognition (ASR),
    then translate with machine translation (MT).
    A major cause of the performance gap is that, while existing AST corpora are small, massive datasets exist for both the ASR and MT subsystems.
    In this work, we evaluate several data augmentation and pretraining approaches for AST, by comparing all on the same datasets.
    Simple data augmentation by translating ASR transcripts proves most effective on the English--French augmented LibriSpeech dataset, 
    closing the performance gap from 8.2 to 1.4 BLEU, compared to a very strong cascade that could directly utilize copious ASR and MT data.
    The same end-to-end approach plus fine-tuning closes the gap on the English--Romanian MuST-C dataset from 6.7 to 3.7 BLEU.
    In addition to these results, we present practical recommendations for augmentation and pretraining approaches. Finally, we decrease the performance gap to 0.01 BLEU using a Transformer-based architecture.

\end{abstract}

\input{introduction.tex}
\input{approach.tex}
\input{models.tex}
\input{experiments.tex}
\input{related_work.tex}
\input{conclusion.tex}

\newpage
\bibliographystyle{IEEEtran}
\bibliography{iwslt2019}

\end{document}

%% file: introduction.tex
\section{Introduction}

Automatic speech-to-text translation (AST) is the task of transforming speech input to its corresponding textual translation. 
Traditionally, AST has been conducted via a cascade approach~\cite{post2013improved,ney1999speech}: an automatic speech recognition (ASR) system creates a transcript of the speech signal, 
which is then translated by a machine translation (MT) system. 
Recent approaches use end-to-end neural network models~\cite{berard2016listen,weiss2017sequence}, which are directly inspired by end-to-end models for ASR~\cite{chorowski2015attention,chan2016listen}. 
Cascade approaches can benefit from large amounts of training data available to its components for certain language pairs. 
For example, an English ASR model can be trained on 960 hours of speech~\cite{panayotov2015librispeech}, and an English--French
translation model can be trained on about 40 million sentence pairs~\cite{bojar-EtAl:2014:W14-33}. 

End-to-end approaches have very limited data available~\cite{kocabiyikoglu2018augmenting,post2013improved,mustc19}; 
nevertheless, they present several benefits. First, end-to-end models can enable lower inference latency since they involve only one prediction. 
Second, it may be easier to reduce the model size for a single integrated model. 
Finally, end-to-end approaches avoid compounding errors from the ASR and MT models.

End-to-end models for AST have been shown to perform better than or on par with cascade models~\cite{weiss2017sequence,berard2018end} when both are trained only on speech translation parallel corpora. 
However, when additional data are used to train its ASR and MT subsystems, the cascade outperforms the vanilla end-to-end approach~\cite{berard2016listen,sperber2019attention,jia2019leveraging}. In this work, we explore several techniques that leverage the wealth of ASR and MT data to aid end-to-end systems, by means of data augmentation.

\begin{figure}[t]
    \centering
    \includegraphics[width=\linewidth]{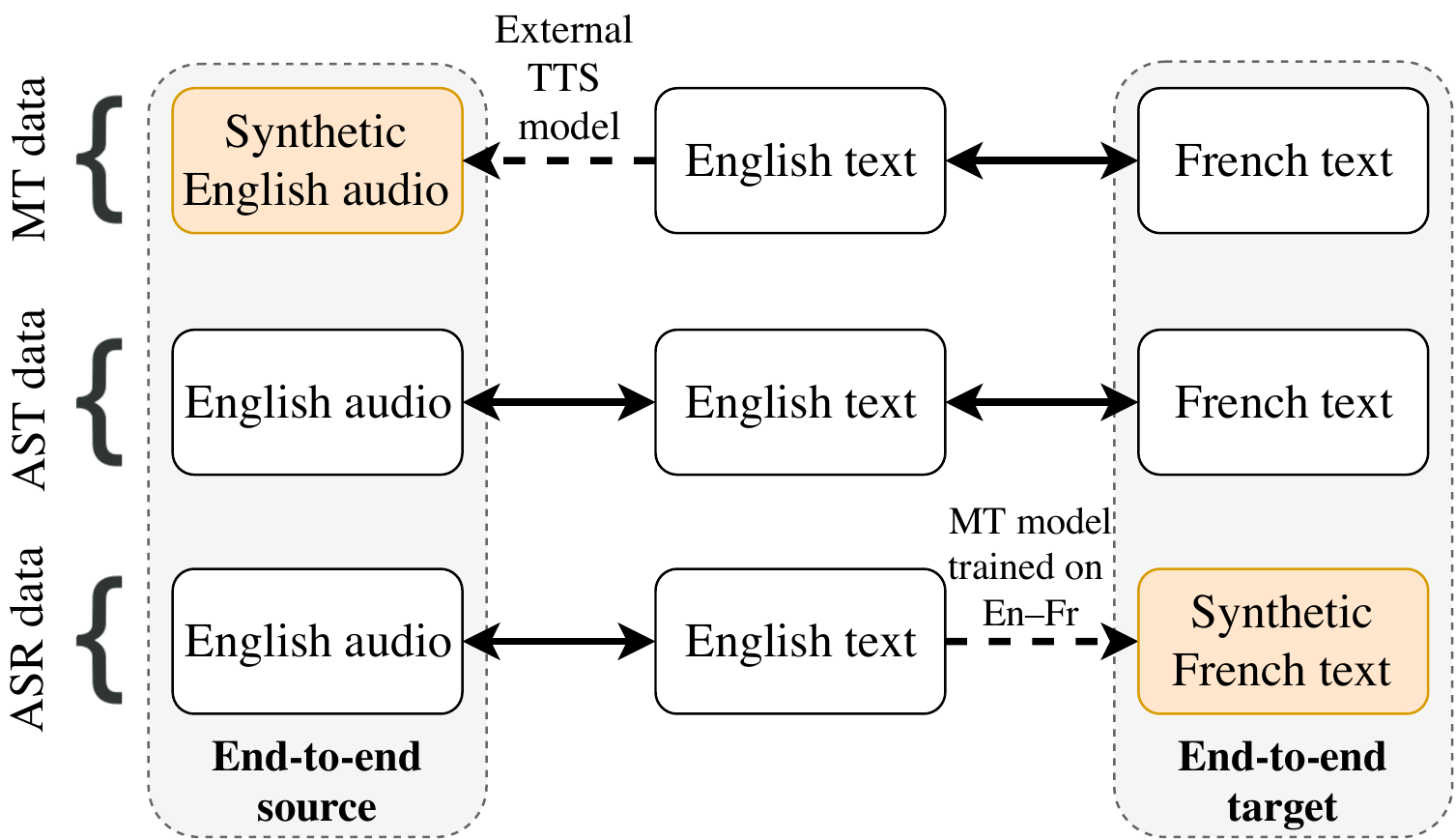}
    \caption{AST datasets typically include three parts: recorded speech, transcripts, and translations. While cascaded models can leverage these, an advantage over end-to-end systems is that they can also leverage data that provide adjacent pairs---which are far more prevalent. The approaches we investigate involve completing the triplets, so end-to-end systems can also benefit from incomplete triplets.}
    \label{fig:dataflow}
\end{figure}

The major contributions of this paper are:
\begin{enumerate}[leftmargin=*]
\item \textbf{We confirm that end-to-end models underperform cascade models by a large margin}, 
when the components can be trained on additional ASR and MT training data while the end-to-end model is constrained to be trained on AST training data. 
In particular, we build a very strong cascade model that outperforms a previously reported system~\cite{berard2018end} by 5.5 BLEU.

\item \textbf{We investigate the strategies that can improve end-to-end AST models}. 
We augment the data by leveraging ASR training data with MT and MT training data with text-to-speech synthesis (TTS). 
We also study the effect of pretraining the ASR encoder as well as how to better utilize out-of-domain augmented data with fine-tuning. 
In the case of TTS-augmented data, we analyze the effect of the amount of data added, which TTS engine is used, and whether one speaker or multiple speakers are used to generate the data.

\item \textbf{We benchmark the performance of several architectures on AST task on public available datasets.}
We first propose an extension to the Bérard model~\cite{berard2018end} that increases its capacity for training on larger data settings. 
We also benchmark models on the AST task that have been previously applied to the ASR task only: VGG LSTM~\cite{watanabe2018espnet} and VGG Transformer~\cite{mohamed2019transformers}. To our knowledge, this is the first time the VGG Transformer architecture has been applied to the AST task.
For better reproducibility, experiments are conducted on two publicly available datasets, AST Librispeech~\cite{kocabiyikoglu2018augmenting} and MuST-C~\cite{mustc19}. 
With data augmentation, pretraining, fine-tuning and careful architecture selection, 
we obtain competitive end-to-end models on the corresponding English--French (En--Fr) and English--Romanian (En--Ro) tasks.

\end{enumerate}


%% file: approach.tex
\section{Approach}
\label{sec:approach}

\begin{figure}[t]
    \centering
    \includegraphics[width=\linewidth]{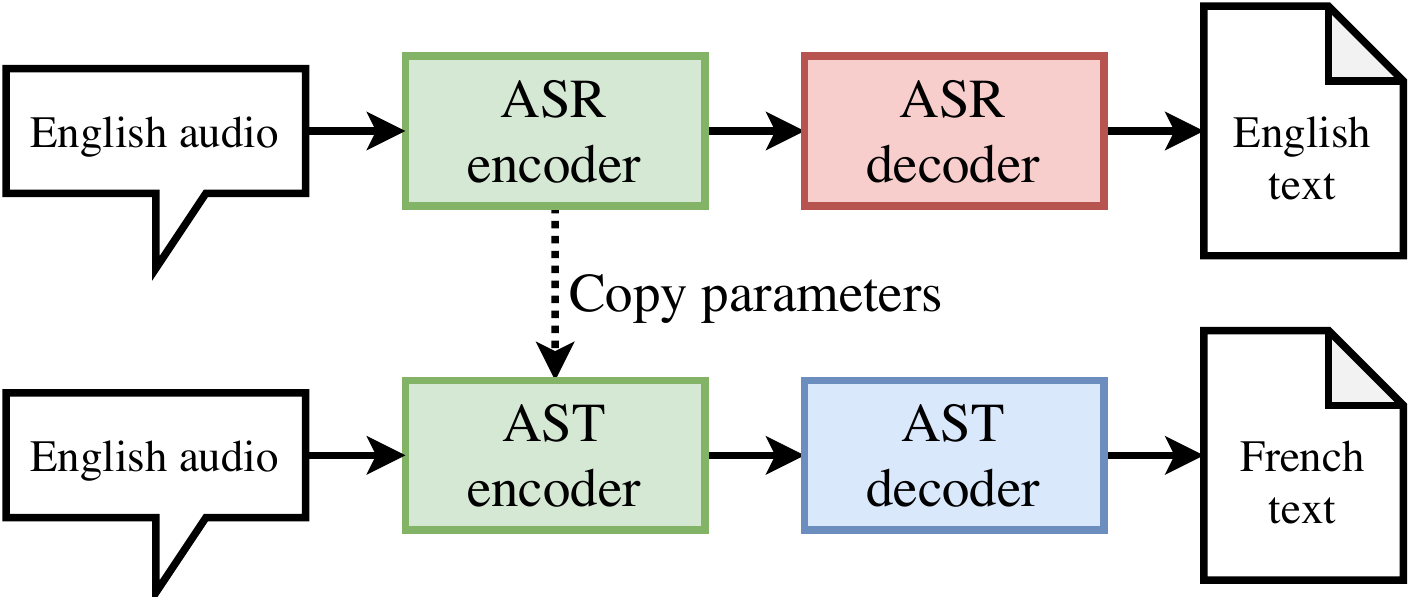}
    \caption{Pre-training end-to-end AST encoder on ASR task.}
    \label{fig:pretraining}
\end{figure}{}

For certain language pairs, cascade models can access large amounts of training data. 
In this section, we present our strategies to leverage this additional data for end-to-end models. 

The first prong of our approach involves generating synthetic data to augment the existing AST data (\autoref{sec:dataaugmentation}), and it is summarised in \autoref{fig:dataflow}.
MT models can be used to generate synthetic AST training data by automatically
translating the transcript portion of ASR training data.
(For our languages of interest, the MT model can be trained on a large amount of data and will be able to generate high quality translations.)
In addition, the MT model itself would not be directly used in training the end-to-end AST model, 
which will avoid reinforcing errors produced by that model. 

Similarly, we can generate additional synthetic AST training data by generating speech from the source side of an MT parallel corpus. 
This technique is similar to backtranslation~\cite{sennrich2015improving} and has been previously applied to end-to-end ASR training~\cite{hayashi2018back}.

Another important aspect to consider is that the additional synthetic data we generate
may not be in-domain. This can be problematic when there is a large gap between the amount of available gold data compared to the amount of synthetic data.
We investigate fine-tuning techniques to address this issue (\autoref{sec:finetuning}).

As a second prong to the approach, we examine pretraining (\autoref{sec:pretraining}): we can use large ASR corpus to pretrain the speech encoder of an AST model; see \autoref{fig:pretraining}.

%% file: models.tex
\section{Models}
\label{sec:models}

In this section, we describe the various model architectures used for ASR and AST experimentation.

\subsection{Bérard Model and Extension}

We use a similar architecture to \cite{berard2018end} with a speech encoder consisting of two non-linear layers 
followed by two convolutional layers and three bidirectional LSTM layers, and a custom LSTM decoder.
In addition, the input state to an LSTM layer is the state emitted at the current timestep by the layer beneath. 
The input state to the bottom layer is the state emitted by the top LSTM layer at the previous timestep. 
Preliminary experiments showed that passing the state from the previous timestep to the LSTM layer one level above at the current timestep was not as effective. 
Finally, we extend the architecture to an arbitrary number $N$ of decoder layers as shown in \autoref{eq:berard}:
\begin{align}
\label{eq:berard}
\begin{split}
    \bm{s}_t^1, \bm{o}_t^1 &= \text{LSTM}^1(\bm{s}_{t-1}^N, \bm{e}(y_{t})) \\
    \bm{c}_t          &= \text{attention}(\bm{o}_t^1, \bm{h}) \\
    \bm{s}_t^n, \bm{o}_t^n &= \text{LSTM}^n(\bm{s}_t^{n - 1}, \bm{c}_t)
\end{split}
\end{align}
\noindent where subscript $t$ indicates the timestep, superscript $n$ indicates the position in the stack of LSTMs, \(\bm{s}\) is the state, 
\(\bm{o}\) is the output, \(\bm{e}\) is the embedding function, \(y\) is a target token and \(\bm{c}\) is a context vector.

\subsection{VGG LSTM}

We investigate the performance of ASR and AST with a model similar to the ESPnet\footnote{\url{https://github.com/espnet/espnet}} implementation~\cite{watanabe2018espnet}. 
The encoder is composed of two blocks of VGG layers~\cite{simonyan2014very} followed by bidirectional LSTM layers. 
We use a hybrid attention mechanism from \cite{chorowski2015attention} that takes into account both location and content information. 
The decoder is an LSTM, following \cite{luong-etal-2015-effective}. This model will subsequently be called \textsc{vgglstm}.

\subsection{VGG Transformer}

We also investigate the performance of a Transformer model. 
To our knowledge, this is the first time this type of model is applied to the AST task. 
We use a variant of the Transformer network ~\cite{vaswani2017attention} that has been shown to perform well on the ASR task
~\cite{mohamed2019transformers} by replacing the sinusoidal positional embedding with input representations learned by convolutions. 
This model will subsequently be called  \textsc{vggtransformer}.

%% file: experiments.tex
\section{Experimental Setup}
\label{sec:experimental_setup}

\subsection{Datasets}
\label{sec:datasets}

For both the En--Fr and En--Ro language pairs, we use three datasets corresponding to the AST, ASR and MT tasks. 
We choose some of the largest publicly available datasets for ASR and MT in order to have the ASR and MT models 
in an unconstrained-like setting and make the comparison between end-to-end and cascade models more realistic. 
Dataset statistics are summarized in \autoref{tab:aststats}.

For the En--Fr AST task, 
we use the publicly available augmented Librispeech corpus~\cite{kocabiyikoglu2018augmenting} (\emph{AST Librispeech}), 
which is the second largest dataset publicly available for this task. We compare our results to the most recent work on this
dataset~\cite{berard2018end}.
For En--Ro AST, we use the recently released MuST-C corpus~\cite{mustc19}, 
which is the largest publicly available dataset, and compare our results to the original dataset benchmarks.

For the En ASR task, we use Librispeech~\cite{panayotov2015librispeech} (\emph{ASR Librispeech}) 
which is in the same domain as AST Librispeech and also the largest publicly available dataset for ASR. 
Since the validation and test set from AST Librispeech come from the training portion of ASR Librispeech, 
we filter all training utterances from ASR Librispeech that contain any of the validation or test utterances from AST Librispeech with at least two words.
We allow ASR Librispeech training utterances to contain validation and test utterances from AST Librispeech with only one word, 
otherwise we would for example remove all training utterances containing the word ``no''. 
The ASR Librispeech corpus is used for the ASR task for both En--Fr and En--Ro experiments.

For the En--Fr MT task, we use the En--Fr parallel data available as part of the WMT14\footnote{\url{http://www.statmt.org/wmt14/translation-task.html}} competition~\cite{bojar-EtAl:2014:W14-33}.
For En--Ro, we use the En--Ro parallel data available for the WMT16\footnote{\url{http://statmt.org/wmt16/translation-task.html}} competition~\cite{bojar-EtAl:2016:WMT1}.
\begin{table}
\small
  \begin{center}
    \begin{tabular}{lrr}
      \toprule
      Dataset         & \# utterances & \# hours \\
      \midrule
      \multicolumn{3}{c}{En--Fr} \\
      \midrule
      AST Librispeech & 94,542 & 100 \\
      ASR Librispeech & 265,754 & 902 \\
      WMT14 & 29,362,441 & - \\
      dev & 1071 & 2 \\
      test & 2048 & 3:44 \\
      \midrule
      \multicolumn{3}{c}{En--Ro} \\
      \midrule
      MuST-C & 236,168 & 432 \\
      WMT16 & 612,422 & - \\
      dev & 1370 & 2:33 \\
      tst-COMMON & 2556 & 4:10 \\
      \bottomrule
    \end{tabular}
    \caption{Dataset statistics for the AST, ASR and MT tasks for En--Fr and En--Ro. The development and test sets are common to the ASR and AST tasks. tst-COMMON is the common test set for MuST-C.}
    \label{tab:aststats}
  \end{center}
\end{table}

\subsection{Preprocessing Settings}

For the En--Fr task, we follow the same preprocessing as \cite{berard2018end}. 
The English text is simply lowercased as it does not contain punctuation.
The French text is punctuation-normalized, tokenized and lowercased. 
For the En--Ro task, the English text is tokenized, punctuation-stripped and lowercased.
The Romanian text is punctuation-normalized and tokenized but the casing is preserved, following \cite{mustc19}.
We do not limit the number of frames in the training data except to avoid GPU out-of-memory errors.

When training on AST Librispeech only, we use a character-level decoder. 
Otherwise, we use a unigram model with size 10,000 using the SentencePiece implementation~\cite{kudo2018sentencepiece} as training on larger datasets with a character-level decoder would be prohibitively slow.

\subsection{Model Settings}
We use two sets of hyperparameters for the Bérard architecture. 
When training on AST Librispeech (En--Fr), we reuse the same parameters as \cite{berard2018end}.
In all other settings---for En--Fr with additional data and En--Ro---we use 3 decoder layers based on the extended model presented in \autoref{sec:models} in order to give more capacity to the model.

The \textsc{vgglstm} encoder uses 80 log-scaled mel spectrogram features, 2 VGG blocks with 64 and 128 channels, filter size 3, 
pooling size 2, 2 convolutional layers and layer normalization and 5 bidirectional LSTM layers of size 1024. 
The decoder uses embeddings of size 1024 and 2 LSTM layers of size 1024. 
The attention has dimension 1024 and 10 channels with filter size 201. 
\textsc{vgglstm} uses no dropout.

Our \textsc{vggtransformer} also uses 80 features, the same VGG block configuration as \textsc{vgglstm}, 
14 transformer encoder layers and 4 transformer decoder layers with size 1024, 16 heads, 
a feed forward network of size 4096 and dropout with probability 0.15. 
The \textsc{vggtransformer} decoder uses target embeddings of size 128 and 4 convolutional layers with 256 channels, filter size 3 and layer normalization.

\autoref{tab:num_params} gives the number of parameters for these four models as well as the Transformer model used for MT.

\begin{table}
  \begin{center}
    \begin{tabular}{lr}
      \toprule
      Architecture & \# Parameters \\
      \midrule
      Bérard      &  8.9M \\
      Bérard with 3 decoder layers & 13.5M \\
      VGG LSTM         & 176M\\
      VGG Transformer    & 258.9M \\
      Transformer (MT)    & 214.2M \\
      \bottomrule
    \end{tabular}
    \caption{Number of parameters for each model architecture.}
    \label{tab:num_params}
  \end{center}
\end{table}

\subsection{Training Settings}

For the Bérard architecture, we use the Adam optimizer~\cite{kingma2014adam} with a learning rate of 0.001. 
For the smaller AST Librispeech task, we use a minibatch size of 16000 frames to help convergence. 
For other tasks, we use a minibatch size of 96,000 frames except for the \textsc{vggtransformer} where we use 72,000 frames (to avoid memory issues). 
We also use delayed updates~\cite{saunders-etal-2018-multi} in order to keep the same effective batch size and avoid GPU out-of-memory errors. 
All experiments are conducted on 8 GPUs.
For other architectures than Bérard, we use ADADELTA~\cite{zeiler2012adadelta} with a learning rate of 1 and we normalize the loss per utterance instead of per token. 
These hyperparameters were chosen based on preliminary experimentation on the ASR Librispeech task.

\section{Experiments}
\label{sec:experiment}

\subsection{Cascade Baselines}
\label{sec:baselines}

The baseline approach, \textsc{cascade}, involves two steps: first, transcribe input speech with an ASR model, 
then translate the transcript with an MT model. 
Both models are trained separately on large training datasets.

The ASR models for En--Fr use the same architectures from
\autoref{sec:models} and are trained on the full Librispeech corpus, 
which is much larger than the available AST data. For the En--Ro task, 
ASR models are trained on the MuST-C and the Librispeech datasets. 
We use a Transformer~\cite{vaswani2017attention} as the basic MT architecture. 
More precisely, for En--Fr, we first pretrain a large Transformer model (\emph{transformer big}) over the entire WMT14 corpus, 
then fine-tune this model on the AST Librispeech data. 
For En--Ro, we merge the MuST-C and the WMT16 corpora since they have comparable sizes and train a smaller Transformer (\emph{transformer base}) on the joint corpus.




\subsection{Data Augmentation} \label{sec:dataaugmentation}
\label{sec:data_augmentation_description}

\subsubsection{MT: Producing AST data from ASR data.} 
The MT models described in \autoref{sec:baselines} are used to automatically translate the English transcript of ASR Librispeech into French and Romanian. 
The resulting synthetic data can directly be used as additional training data for end-to-end AST models.

\subsubsection{TTS: Producing AST data from MT data.} 
We also explore augmenting the MT training data with TTS. 
This technique is similar to backtranslation~\cite{sennrich2015improving} and has been previously applied to end-to-end ASR training~\cite{hayashi2018back}. 
We use two pretrained TTS engines. 
\textbf{TTS1} uses the OpenSeq2Seq framework~\cite{openseq2seq} to generate speech samples in five different voices. 
The TTS model is based on an extension of the Tacotron 2 model~\cite{shen2018natural} with Global Style Tokens~\cite{wang2018style}. 
\textbf{TTS2} is trained on about 15 hours of single speaker data. 
The text comes from several domains such as Wikipedia, news articles, parliament speech, and novels. 
We use TTS1 to generate speech from a random sample of WMT14 with the same size as ASR Librispeech (265,754 utterances) and TTS2 to generate speech from WMT16 (612,422 utterances).

\subsection{Speech Encoder Pretraining} \label{sec:pretraining}
Speech encoder pretraining is another way to use the full ASR Librispeech dataset \cite{bansal2018pre}.
We first pretrain an English ASR model on ASR Libirspeech plus the TTS1 corpus generated in \autoref{sec:data_augmentation_description}---the parallel corpus built from the generated TTS and WMT14 English text. 
We then take the encoder of the ASR model to initialize the encoder of an AST model with the same architecture. 




\subsection{Fine-tuning} \label{sec:finetuning}
On the En--Fr task, the TTS data is generated from WMT14, which is out-of-domain with respect to Librispeech. 
On the En--Ro task, both the TTS and the MT data are out-of-domain. 
We investigate fine-tuning as a technique to mitigate the domain shift. 
We fine-tune by continuing training on only AST Librispeech or MuST-C, starting from the best checkpoint on development data after convergence.

\section{Results and Analysis}
\label{sec:results}
\begin{table}[t]
\centering
\begin{adjustbox}{max width=\linewidth}
\begin{tabular}[b]{l l l l l}
\toprule
Task & Model & Data & En--Fr & En--Ro\\
\midrule
\multirow{2}{*}{ASR}  & Bérard \cite{berard2018end} & AST & \metric{15.1}\% & \metric{27.61}\% \\
& Bérard     & ASR & \metric{9.97618}\% &\metric{19.72156}\% \\
\midrule
\multirow{2}{*}{MT}   & Bérard \cite{berard2018end} & AST & \metric{19.3} &\metric{28.16}  \\
& Transformer          & AST + MT & \metric{24.69} &\metric{28.7} \\
\midrule
\multirow{12}{*}{AST}
& oracle BLEU \cite{berard2018end} & --- &  \metric{19.3} \\
& Cascade \cite{berard2018end} &  ASR + AST + MT & \metric{15.8} &  \\
& Our cascade &  ASR + AST + MT & \metric{21.31} &\metric{21.0} \\
& Bérard & AST & \metric{13.1} & \metric{14.3} \\
\cmidrule{2-5}
& Bérard & \multirow{2}{*}{AST + MT} & \metric{19.87345} &\metric{16.42994284149}\\
& + pretraining &  & \metric{19.16647} & \metric{17.320667284412} \\
\cmidrule{2-5}
& Bérard & \multirow{4}{*}{AST + TTS} & \metric{12.589218661071} & \metric{10.34052}\\
& + fine-tuning &  & \metric{14.343967996113} & \metric{12.519535700313} \\
& + pretraining &  & \metric{15.19753} & \metric{13.042834798466}\\
& + pretraining + fine-tuning &  & \metric{16.350647768811} & \metric{15.204942876671} \\
\cmidrule{2-5}
& Bérard & \multirow{4}{*}{AST + MT + TTS} & \metric{19.04978477461} & \metric{10.025473524292}\\
& + fine-tuning &  & \metric{19.294478180314} & \metric{12.287110632688}\\
& + pretraining &  & \metric{19.190454410913} & \metric{14.25468761405}\\
& + pretraining + fine-tuning &  & \metric{19.224063906248} & \metric{15.653703459845}\\
\bottomrule
\end{tabular}
\end{adjustbox}
\caption{
  En--Fr results on AST Librispeech test set and En--Ro results on MuST-C test set with Bérard model. 
  For ASR, we use the same AST subset of Librispeech as the ASR test set. 
  ASR performance is measured by word error rate (WER; lower is better) and MT and AST performance is measured by BLEU (higher is better).
  MT, TTS, fine-tuning and pretraining help bridge the gap to a very strong cascade model.}
\label{tab:bigtable-1}
\end{table}

We first investigate techniques to improve end-to-end AST on Bérard model.
Results for the En--Fr and En--Ro tasks are summarized in \autoref{tab:bigtable-1}.
The ASR and MT components we trained with additional data are very strong: the cascade model outperforms the vanilla end-to-end model by 8.2 BLEU (21.3 vs.\ 13.1) on the AST task.
It is also important to note that our cascade baseline is greater than the best reported result on this task by 5.5 BLEU 
and even better than the previously reported oracle BLEU of 19.3 by 2 BLEU, where the gold transcript is passed to the translation system.

\subsection{Effect of Data Augmentation}

By augmenting ASR Librispeech with automatic translations (\textbf{AST + MT}), 
we show an improvement of 6.8 BLEU for En--Fr and 3.0 BLEU for En--Ro. 
Under this setting, we observe that pretraining is not beneficial for En--Fr but is  for En--Ro.\footnote{Data setting experiments were conducted on the B\'{e}rard architecture. We do not expect the conclusions on data augmentation, pretraining and fine-tuning to change on the VGG architectures.} In this setting on En--Ro, we close the gap between cascade and end-to-end BLEU from 6.7 to 3.7.

Additional TTS-augmented data (\textbf{AST + TTS}) initially hurts performance for both language pairs. 
With pretraining and fine-tuning, TTS data provides a gain of 3.3 BLEU over the vanilla end-to-end baseline. 
However, it still underperforms the model using MT-augmented data only.

We also augmented the data with MT and TTS (\textbf{AST + MT + TTS}) at the same time. We find that it does not provide additional gain over using MT data only. 
In general, MT data can efficiently help the model, while TTS data is less efficient and can be hurtful. 
We analyze TTS in more detail in \autoref{subsection:tts-data}.

\begin{table}[t]
\small
  \centering
  \begin{tabular}[b]{l r}
\toprule
  Architecture & BLEU \\
\midrule
        \multicolumn{2}{c}{Cascade (ASR architecture; Transformer used for MT)} \\
\midrule
  \textsc{bérard} & \metric{21.31} \\
  \textsc{vgglstm} & \metric{21.79} \\
 \textsc{vggtransformer} & \metric{21.66} \\
\midrule
        \multicolumn{2}{c}{End-to-end AST} \\
\midrule
  \textsc{bérard} & \metric{19.87} \\
  \textsc{vgglstm} & \metric{19.84} \\
\textsc{vggtransformer} & \metric{21.65} \\
\bottomrule
\end{tabular}
  \caption{
    Performance of the architectures we consider on the (En--Fr) AST Librispeech test set in the AST + MT data setting, which performed best in \autoref{tab:bigtable-1}. The large \textsc{vggtransformer} outperforms other end-to-end approaches, now within 0.15 BLEU of the best cascade model.
   }
  \label{tab:arch-en-fr}
  \end{table}
We present the performance of different \textbf{architectures} on the higher-resource En--Fr task in \autoref{tab:arch-en-fr}. In the AST + MT setting, 
we obtain BLEU of \metric{19.87} with Bérard, \metric{19.84} with \textsc{vgglstm} and \metric{21.65} with \textsc{vggtransformer}, showing that these architectures are effective with additional data.
\textsc{vggtransformer} in particular (obtaining the best end-to-end AST score) 
is on par with the \textsc{vggtransformer} cascade and \metric{-0.1} BLEU behind the \textsc{vgglstm} cascade.

\subsection{Effect of Encoder Pretraining}

Results are summarized in \autoref{tab:pretraining}.
Pretraining improves the AST + TTS system by +2.6 BLEU and +2.0 BLEU with fine-tuning, improves the MuST-C + TTS system by +2.7 BLEU and +2.2 BLEU with fine-tuning, and improves the MuST-C + MT + TTS system by +4.2 BLEU and +1.4 BLEU. 

However, gains from pretraining ASR on the full Librispeech dataset do not compound with gains from MT augmentation. Pretraining does not help as much in the AST + MT + TTS setup, showing a negligble change in BLEU score. Pretraining has mixed results for the MuST-C + MT case, with -0.2 BLEU and +0.8 BLEU with pretraining.

Pretraining on in-domain ASR data is not a good substitute for MT-augmenting the ASR data. However we note that using a pretrained speech encoder will speed up convergence of the AST model. Thus, pretraining could be used in experiments with the same architecture and provides a good starting point for more rapid iteration.


\begin{table}
  \small
  \begin{center}
    \begin{tabular}{l >{\raggedleft\arraybackslash}p{0.9cm} >{\raggedleft\arraybackslash}p{0.9cm} >{\raggedleft\arraybackslash}p{0.9cm} >{\raggedleft\arraybackslash}p{0.9cm}}
      \toprule
       Dataset & Bérard & $\Delta$ PT &  + FT & $\Delta$ PT \\
      \midrule
      AST + TTS & \metric{12.58922} & \metric{2.60831} & \metric{14.34397} & \metric{2.00668} \\
      AST + MT + TTS  & \metric{19.04978} & \metric{0.14067} & \metric{19.19045} & \metric{0.03361} \\
      MuST-C + TTS & \metric{10.34052} & \metric{2.70231} & \metric{13.04283} & \metric{2.16211} \\
      MuST-C + MT & \metric{16.42994} & \(-\)\metric{0.21588} & \metric{15.71237} & \metric{0.82949} \\
      MuST-C + MT + TTS & \metric{10.02547} & \metric{4.22921} & \metric{14.25469} & \metric{1.39902} \\
      \bottomrule
    \end{tabular}
    \caption{Effect of pretraining (PT), when added to the baseline or the fine-tuned (FT) baseline.}
    \label{tab:pretraining}
  \end{center}
  \vspace{-10pt}
\end{table}

\subsection{Effect of Fine-tuning}

\autoref{tab:finetuningresults} summarizes the fine-tuning results.
We apply fine-tuning whenever TTS-augmentation is used. Fine-tuning seems to mitigate the effect of domain shift introduced by the additional out-of-domain TTS data: in the Librispeech AST + TTS setup, fine-tuning improves by +1.7 (+1.2 BLEU on top of pretraining). For the MuST-C + MT setup, we see +2.3 BLEU (+1.4 on top of pretraining).  

Fine-tuning does not improve the AST Librispeech model on top of MT-augmentation though, likely because the MT-augmented data is already in-domain. For the AST + TTS + MT setup, we see neutral results: +0.3 BLEU and no effect on top of pretraining. However, for the MuST-C + MT setup we see a gain of +0.3--0.9 BLEU because the MT-augmented data is out-of-domain for the MuST-C dataset.

\begin{table}
  \small
  \begin{center}
    \begin{tabular}{l p{0.9cm} p{0.9cm} p{0.9cm} p{0.9cm}}
      \toprule
       Dataset & Bérard & $\Delta$ FT & + preT & $\Delta$ FT \\
      \midrule
      AST + TTS & \metric{12.58922} & \metric{1.75475} & \metric{15.19753} & \metric{1.15312} \\
      AST + MT + TTS  & \metric{19.04978} & \metric{0.24469} & \metric{19.19045} & \metric{0.03361} \\
      MuST-C + TTS & \metric{10.34052} & \metric{2.17902} & \metric{13.04283} & \metric{2.16211} \\
      MuST-C + MT & \metric{16.42994} & \metric{0.89072} & \metric{16.21407} & \metric{0.32779} \\
      MuST-C + MT + TTS & \metric{10.02547} & \metric{2.26164} & \metric{14.25469} & \metric{1.39902} \\
      \bottomrule
    \end{tabular}
    \caption{Effect of fine-tuning (FT), both when added to the baseline or the pretrained (preT) baseline.}
    \label{tab:finetuningresults}
  \end{center}
\end{table}



\subsection{TTS Data: Quantity, Quality and Diversity}
\label{subsection:tts-data}

\begin{figure}
\begin{center}
  \includegraphics[width=0.7\linewidth]{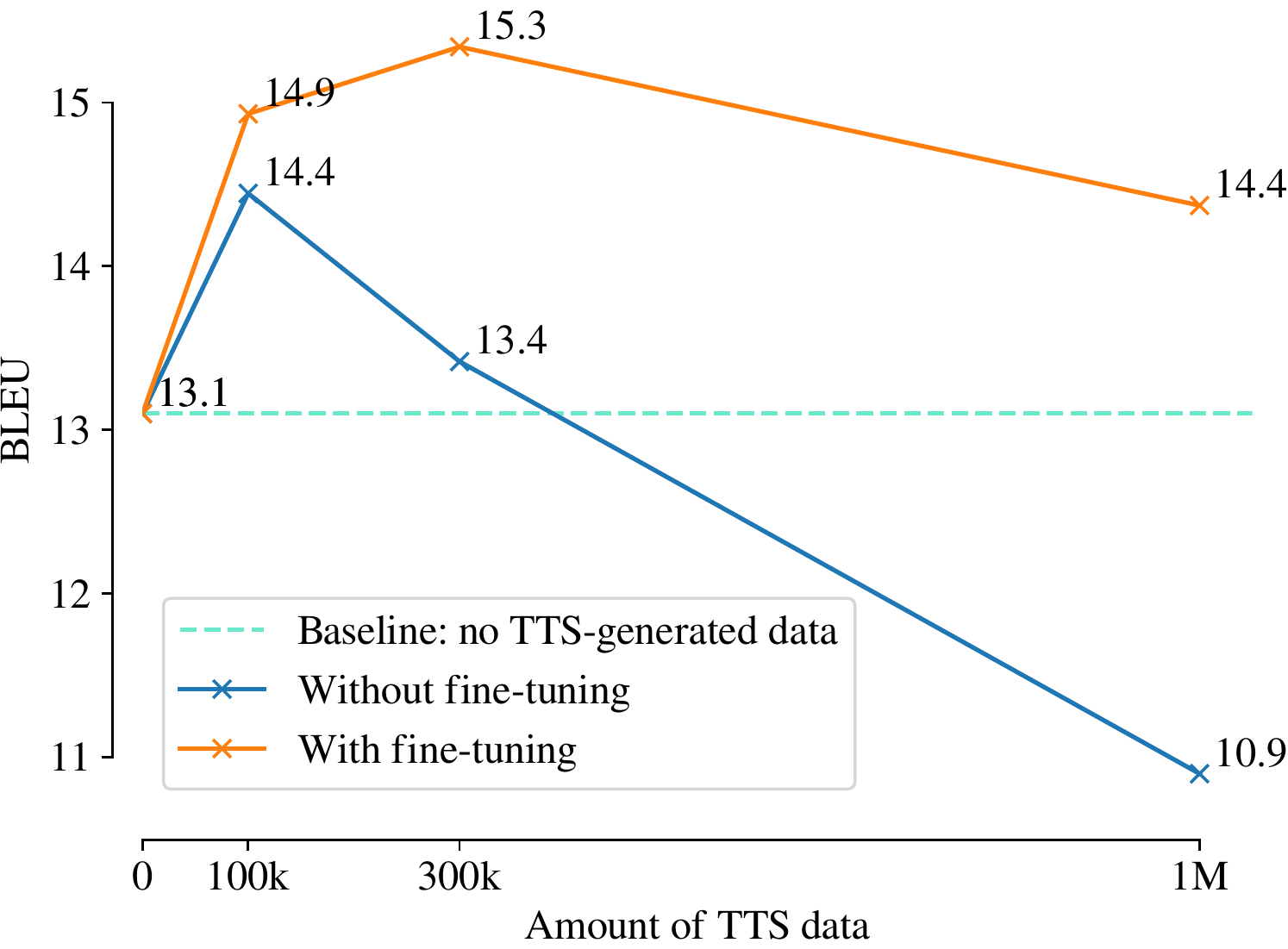}
\caption{Varying amounts of TTS data for the En--Fr task, with and without fine-tuning.}
\label{fig:tts_amount}
\end{center}
\end{figure}

How does augmenting the AST training data with TTS affect performance in the En--Fr task?
%
First, in \autoref{fig:tts_amount}, we see that while adding TTS data up to 100,000 utterances improves the performance, the performance degrades beyond that.
We hypothesize that this is because the additional TTS data is out of domain. With fine-tuning, adding up to 300,000 utterances improves performance, which confirms our hypothesis; however, adding 1M utterances starts degrading performance. In the future, we will investigate how to make more effective use of larger quantities of TTS-generated data.

We also study the effect of using single-speaker or multi-speaker TTS. We use a sample of size 300k utterances from WMT14 and generate speech with the TTS1 engine using the first speaker Speaker 0, the second speaker Speaker 1, and all five speakers in a round-robin fashion. Finally, we investigate whether the quality of the TTS engine matters. For the same sample of 300k utterances, we generate speech using the TTS2 engine both from the English text and the corresponding French translations. The latter is analogous to copying the target to the source in machine translation~\cite{currey-etal-2017-copied}. 

Results are reported in \autoref{tab:tts_speaker_engine}. Comparing the first two rows, we conclude that
performance may vary depending on the speaker. The third row
shows that using multiple speakers performs on par with choosing Speaker 0 (+0.2 BLEU) but outperforms choosing Speaker 1 (+0.9 BLEU). In the future, we will investigate whether the multi speaker approach can mitigate the effect observed in \autoref{fig:tts_amount}.
Finally, comparing rows 1 and 4, we conclude that the quality of the TTS may
matter marginally, with TTS2 slightly outperforming TTS1. The last row shows that
the analogue of copying target to the source in machine translation is an interesting avenue fo further investigation.

\begin{table}
  \small
  \begin{center}
    \begin{tabular}{rrrr}
      \toprule
      Row \# & Configuration & TTS Engine & BLEU \\
      \midrule
      1 & Speaker 0 & TTS1 & \metric{13.419145678342} \\
      2 & Speaker 1 & TTS1  & \metric{12.670337489026} \\
      3 & Multi speaker & TTS1 & \metric{13.570877802067} \\
      4 & English & TTS2 & \metric{13.573597558123} \\
      5 & French & TTS2 & \metric{13.646334127867} \\
      \bottomrule
    \end{tabular}
    \caption{Effect of the number of speakers and the TTS engine when adding TTS data.}
    \label{tab:tts_speaker_engine}
  \end{center}
\end{table}


%% file: related_work.tex
\section{Related Work}
\label{sec:related_work}

Initial attempts at speech translation \cite{ney1999speech} incorporate lattices from ASR systems as inputs to statistical MT models \cite{matusov2005integration,casacuberta2008recent}.
More recent approaches have focused on end-to-end models. \citet{berard2016listen} demonstrate the viability of this approach on a
small synthetic corpus. \citet{weiss2017sequence} outperform a cascade model using a similar architecture to an attention-based
ASR model. \citet{weiss2017sequence} and \citet{berard2018end} show that multi-task learning can further improve an end-to-end model.
Pretraining has also been shown to improve end-to-end models~\cite{berard2018end,bansal2018pre}.

\citet{sperber2019attention} note that cascaded models are at a disadvantage when constrained to be trained on speech translation data only. They show how to leverage additional ASR and MT training data with an attention-passing mechanism. \citet{jia2019leveraging} improve an end-to-end model using MT-augmented and TTS-augmented data but do so \emph{on a proprietary dataset}. In contrast, we experiment on two public datasets where we obtain new state-of-the-art performance; further, we provide additional analyses on network architectures and recommendations on how to better leverage TTS-augmented data.



%% file: conclusion.tex
\section{Conclusion}
\label{sec:conclusion}

We have demonstrated that cascaded models are very competitive when not constrained to only train on AST data. We have studied several techniques aimed at bridging the gap between end-to-end and cascade models.
With data augmentation, pretraining, fine-tuning and architecture selection, we trained end-to-end models that show competitive performance when compared to cascade approach. Our approaches reduced the performance gap between end-to-end and strong cascade models, from 8.2 to 1.4 BLEU on En--Fr Librispeech AST data and from 6.7 to 3.7 on the En--Ro MuST-C corpus.
We also analyzed the effect of TTS data in terms of quality, quantity, and the use of single speaker vs.\ multiple speakers, and we provide recommendations on how to harness this type of data. In the future, we would like to investigate how to better use larger-scale TTS-generated data.